\documentclass[runningheads]{llncs}

\usepackage[mobile]{eccv}

\usepackage{eccvabbrv}

\usepackage{graphicx}
\usepackage{booktabs}
\usepackage{algorithm}
\usepackage[noend]{algpseudocode}

\usepackage[accsupp]{axessibility}  

\usepackage[pagebackref,breaklinks,colorlinks,citecolor=eccvblue]{hyperref}

\usepackage{orcidlink}

\begin{document}

\title{Deep Reward Supervisions for Tuning Text-to-Image Diffusion Models} 

\titlerunning{Deep Reward Supervisions for Tuning Text-to-Image Diffusion Models}

\author{Xiaoshi Wu*\inst{1,3} \and
Yiming Hao*\inst{2} \and
Manyuan Zhang\inst{1} \and Keqiang Sun\inst{1} \and \\ Zhaoyang Huang\inst{3} \and Guanglu Song\inst{4} \and Yu Liu\inst{4} \and Hongsheng Li\inst{1,2}}

\authorrunning{Wu \etal}

\institute{Multimedia Laboratory, The Chinese University of Hong Kong \\
 \and Centre for Perceptual and Interactive Intelligence (CPII) \\ \and Avolution AI \and SenseTime  \\
\email{wuxiaoshi@link.cuhk.edu.hk}}

\maketitle

\vspace{-1cm}
\begin{abstract}
Optimizing a text-to-image diffusion model with a given reward function is an important but underexplored research area.
In this study, we propose \textbf{Deep Reward Tuning} (DRTune), an algorithm that directly supervises the final output image of a text-to-image diffusion model and back-propagates through the iterative sampling process to the input noise.
We find that training earlier steps in the sampling process is crucial for low-level rewards, and deep supervision can be achieved efficiently and effectively by stopping the gradient of the denoising network input.
DRTune is extensively evaluated on various reward models.
It consistently outperforms other algorithms, particularly for low-level control signals, where all shallow supervision methods fail.
Additionally, we fine-tune Stable Diffusion XL 1.0 (SDXL 1.0) model via DRTune to optimize Human Preference Score v2.1, resulting in the Favorable Diffusion XL 1.0 (FDXL 1.0) model. FDXL 1.0 significantly enhances image quality compared to SDXL 1.0 and reaches comparable quality compared with Midjourney v5.2. 
\footnote{Authors with * contributed equally to this work.}
\end{abstract}

\begin{center}
    \centering
    \captionsetup{type=figure}
    \includegraphics[width=0.85\textwidth]{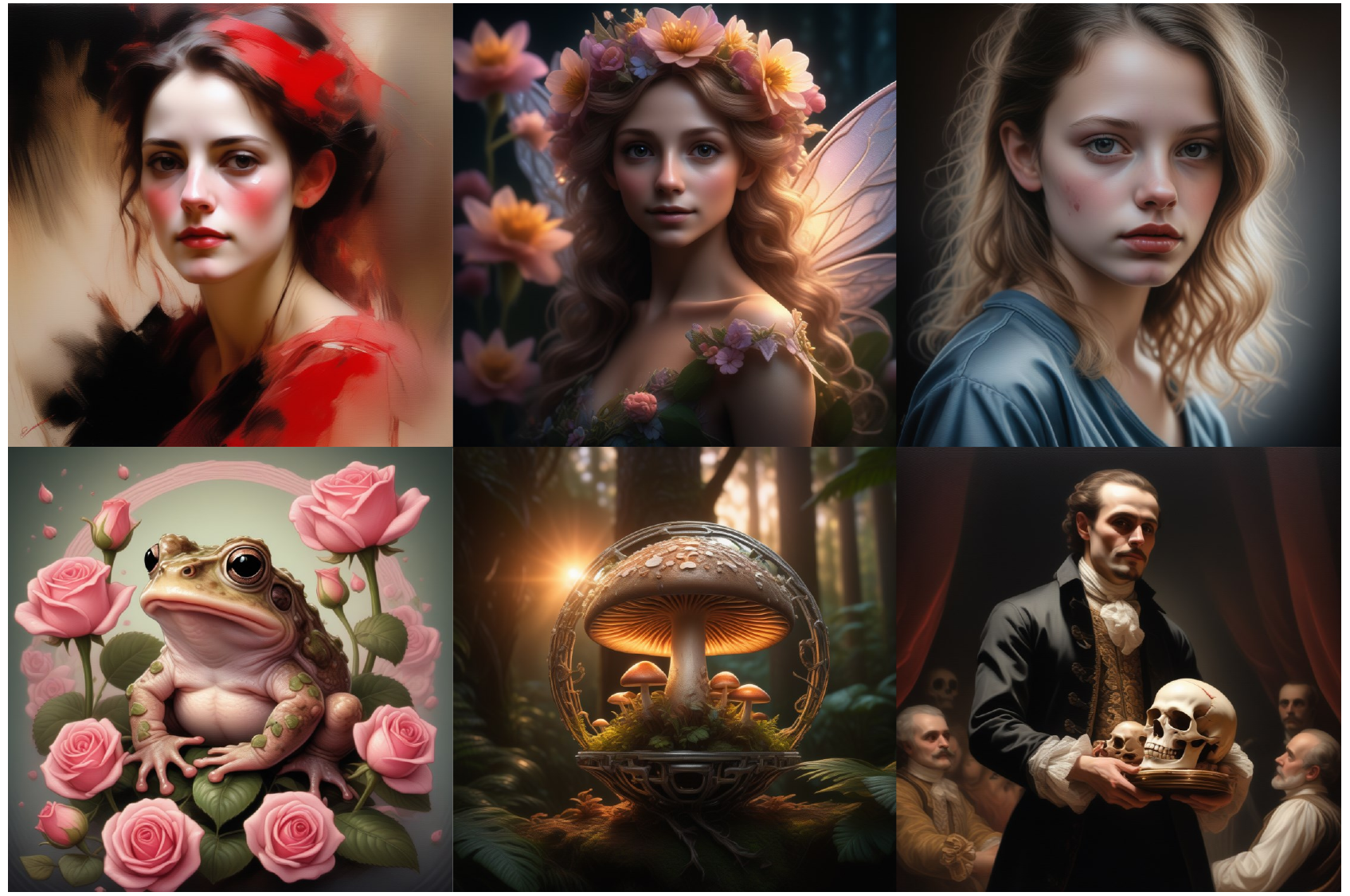}
    \captionof{figure}{Images generated by Favorable Diffusion XL 1.0 (FDXL 1.0). FDXL 1.0 is initialized from Stable Diffusion XL 1.0 (SDXL 1.0) and then trained to optimize Human Preference Score v2.1 via the proposed Deep Reward Tuning.}
\end{center}%

\section{Introduction}
Diffusion models~\cite{sohl2015deep, ho2020denoising} are generative models that sample data from a certain distribution by iteratively denoising from random input, and have been proven effective for image creation~\cite{sohl2015deep, dalle2, stable_diffusion, imagen, podell2023sdxl, ruiz2023dreambooth, zhang2023adding}. However, the iterative denoising paradigm makes it less straightforward to train a diffusion model with gradients from a reward model, compared with generative adversarial networks (GANs)~\cite{goodfellow2020generative}.
Training a diffusion model to optimize a given reward is especially useful in cases where the training target can not be easily characterized by a set of images, such as compressibility and symmetry.

Inspired by \textbf{R}einforcement \textbf{L}earning from \textbf{H}uman \textbf{F}eedback (RLHF)~\cite{ziegler2019fine, stiennon2020learning, ouyang2022training, bai2022training} in natural language processing (NLP), previous works like ReFL~\cite{xu2023imagereward}, DRaFT~\cite{clark2023directly}, and AlignProp~\cite{prabhudesai2023aligning} explore directly supervising the final output of a diffusion model by a differentiable reward function. However, text-to-image diffusion models typically require more than 20 sampling steps~\cite{ddim, dpm, dpm++}. Directly back-propagating step-by-step from the output image to the input noise results in significant memory consumption, as discussed in AlignProp~\cite{prabhudesai2023aligning}. 
To avoid this issue, ReFL~\cite{xu2023imagereward} and DRaFT~\cite{clark2023directly} ignore earlier sampling steps and only train the last few steps before the output image.
However, we find that this strategy is suboptimal, and it fails to optimize certain low-level reward, such as rewards encouraging symmetry images.
We call this a depth-efficiency dilemma.

In this work, we propose \textbf{DRTune} as a solution to the depth-efficiency dilemma. In DRTune, two main modifications are made comparing to the naive step-by-step back-propagation approach: 1) we stop the gradients of the denoising network input, and 2) in each training iteration, we sample a subset of equally spaced steps among all $T$ steps in the sampling process for back-propagation. This design has two advantages: 1) it skips back-propagation for untrained intermediate steps, and enables the optimization of early denoising steps without huge memory consumption and computation.
2) the stop gradient operation solves the gradient explosion issue~\cite{clark2023directly} when training early timesteps, which significantly accelerates convergence. 

We evaluate DRTune by comparing it with baseline methods on a benchmark of 7 rewards, and DRTune consistently achieves higher rewards given the same computation budget, demonstrating the effectiveness of DRTune. Next, we fine-tune Stable Diffusion XL 1.0 (SDXL 1.0) using HPS v2.1~\cite{wu2023humanv2}, producing Favorable Diffusion XL 1.0 (FDXL 1.0), which exhibits significantly better visual quality compared to the base model SDXL 1.0 and even achieves comparable quality with Midjourney v5.2.

The contributions of this work can be summarized as:
1) We propose DRTune for efficiently and effectively supervising early denoising steps of a diffusion model.
2) We introduce FDXL 1.0, the state-of-the-art open-source text-to-image generative model tuned on human preferences.

\section{Related Works}

\noindent\textbf{Reward-training for diffusion models.}
Optimizing text-to-image generative models given a reward function is crucial for learning properties that are difficult to define using a set of images, such as human preference~\cite{wu2023human, xu2023imagereward, kirstain2023pick}, 3D consistency~\cite{sun2022controllable, sun2022cgof++, piao2021inverting}, and additional control~\cite{zhang2023adding}.
Back-propagating through the sampling process of diffusion models has been the focus of several research studies. DiffusionCLIP~\cite{kim2022diffusionclip} was the first to explore fine-tuning a diffusion model for image manipulation. Watson \etal\cite{watson2021learning} adopted a similar technique, but optimize sampler parameters. DOODL\cite{wallace2023end} focused on optimizing the input noise to enhance classifier guidance. In addition, Lee \etal\cite{lee2023aligning}, Wu \etal\cite{wu2023better}, DPOK~\cite{fan2023dpok}, and DDPO~\cite{black2023training} explored fine-tuning text-to-image diffusion models using reinforcement learning to optimize rewards.
An advantage of these methods is that they do not require  rewards to be differentiable, which is beneficial when dealing with non-differentiable reward functions. However, these approaches lead to slower convergence compared to methods exploiting the gradients from rewards.
Recent works, such as DRaFT~\cite{clark2023directly} and AlignProp~\cite{prabhudesai2023aligning}, focuses on optimizing diffusion models using differentiable rewards of human preference~\cite{kirstain2023pick, wu2023humanv2}, which effectively improve the image quality.
In this work, we identify that there is an unsolved depth-efficiency dilemma in previous methods, and propose our solution.

\noindent\textbf{Rewards for text-to-image diffusion models.}
In fact, any perception model that takes an image as input, and make a prediction can serve as a reward model.
Common reward models for tuning a text-to-image diffusion model are CLIP Score for text-to-image alignment~\cite{clip, kim2022diffusionclip, clark2023directly, lee2023aligning}, human preference~\cite{kirstain2023pick, wu2023humanv2, clark2023directly, wu2023better, prabhudesai2023aligning},
JPEG compressibility~\cite{black2023training, clark2023directly}.
In this work, we explore a new kind of reward of symmetry. Although it is a relatively low-level reward compared with other rewards, previous methods fail to optimize it.
We identify that the key to successfully optimize the symmetry reward is to adopt deep supervision.

\noindent\textbf{Stop gradient for iterative refinement.}
Training a network with iterative refinement is common in a variety of computer vision tasks. It typically involves the formulation:
\begin{equation}
x_{next} = x_{prev} + R(x_{prev}),
\label{eq:refine}
\end{equation}
where $x_{\text{prev}}$ represents the current output and $R$ is a network that refines the current output by predicting a residue. This approach is utilized in two-stage object detectors such as Faster R-CNN~\cite{ren2015faster}, where the box coordinates predicted in the first stage are detached from gradients of the second stage refinement. Similarly, state-of-the-art transformer-based detectors like Deformable DETR~\cite{zhu2020deformable} and DINO~\cite{zhang2022dino} employ intermediate supervision for each decoder output. Optical flow estimation models like RAFT~\cite{teed2020raft} also iteratively refine a warping field following the same formulation. In these cases, the inputs of $R$ are all detached. Given that diffusion models also adhere to this formulation during the inference process, the stop gradient technique may prove effective for optimizing them.

\section{Methods}
In this study, we focus on supervising a text-to-image diffusion model using a differentiable reward model. 
We begin by presenting the background in Sec.\ref{sec:background}, and then introduce our method, Deep Reward Tuning, in Sec.~\ref{sec:DRTune}. 

\begin{figure}
  \centering
  \includegraphics[width=1.0\linewidth]{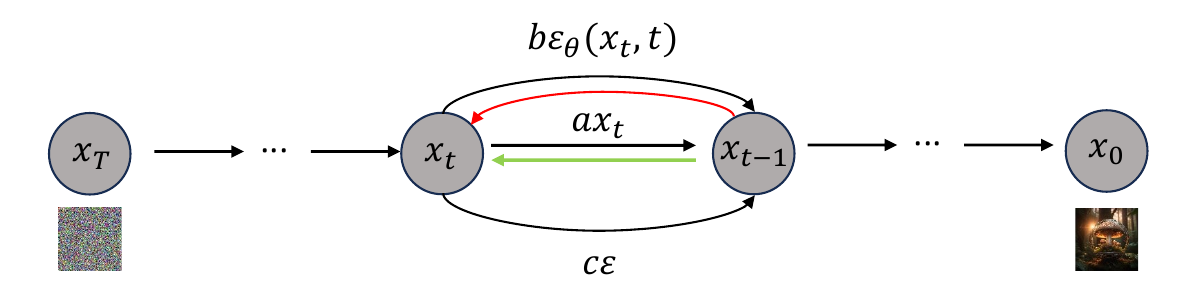}
  \caption{Illustration of the iterative denoising sampling pipeline. When trained, gradients flow through the green branch and the red branch.}
  \label{fig:sampling_illustration}
\end{figure}

\subsection{Background}
\label{sec:background}
\noindent\textbf{Sampling of diffusion models.} The sampling process of a diffusion model is conducted as an iterative denoising procedure. Since this study does not focus on a specific sampling algorithm, we adopt a notation for an abstracted sampler:
\begin{equation}
\mathbf{x}_{t-1} = a_t \mathbf{x}_{t} + b_t \epsilon_{\theta}(\mathbf{x}_t, t) + c_t \epsilon.
\label{eq:sampling}
\end{equation}
At each time step $t$, the denoising model $\epsilon_\theta$ estimates the noise based on the current noisy input $\mathbf{x}_t$ and $t$ to predict the direction towards $\mathbf{x}_0$. Extra conditions for $\epsilon_\theta$ are omitted without loss of generality. $\epsilon\sim\mathcal{N}(0, 1)$ represents Gaussian noise with a mean of 0 and a variance of 1, introducing randomness into the sampling process. The coefficients $a_t, b_t, c_t$ are determined by sampling algorithms and noise schedules. Popular schedulers, such as DDPM~\cite{ho2020denoising}, DDIM~\cite{ddim}, DPM~\cite{dpm, dpm++}, can be parameterized into this framework. In Fig.~\ref{fig:sampling_illustration}, the sampling process is unfolded for better illustration.

\noindent\textbf{Challenges.}
When fine-tuning a diffusion model with gradient on the output image $\mathbf{\hat{x}}_0$, a dilemma arises regarding whether the earlier sampling steps should be trained. Tuning the earlier sampling steps requires back-propagation through the later steps, leading to a significant computation overhead. A previous work~\cite{clark2023directly} has shown that tuning earlier sampling steps can also cause the gradient explosion problem, which hinders convergence. However, given that every single sampling step contributes to the final output, training only the last few steps may not suffice.

\subsection{Deep Reward Tuning}
\label{sec:DRTune}
To address the above challenge, we propose DRTune, an algorithm that can supervise early sampling steps efficiently.
The key idea of DRTune is to block the gradients of denoising network inputs and train a subset of all sampling steps.

\noindent\textbf{Stop gradient.} In DRTune, we address the convergence issue by blocking the gradient of the input $\mathbf{x}_t$:
\begin{equation}
\textcolor{green}{\mathbf{x}_{t-1}} = a_t \textcolor{green}{\mathbf{x}_{t}} + b_t \epsilon_{\theta}(\textcolor{red}{sg}(\mathbf{x}_t), t) + c_t \epsilon,
\label{eq:sampling_bp}
\end{equation}
where $sg(\cdot)$ denotes the stop gradient operation. With this modification, the gradient will only flow through the green linear branch, and the gradients of early steps can be easily computed by multiplying a scalar. We observe that this operation helps alleviate the gradient explosion issue and accelerates convergence.

\noindent\textbf{Training a subset of sampling steps.} After blocking the gradient on the input $\mathbf{x}_t$, the gradients between neighboring time steps satisfy
\begin{equation}
\partial \mathbf{x}_{t-1} = a_t \partial \mathbf{x}_{t}.
\end{equation}
Consequently, the gradient of $\mathbf{x}_t$ can be calculated as
\begin{equation}
\partial \mathbf{x}_{t} = \prod_{s=1}^t a_s^{-1} \partial \mathbf{x}_0.
\end{equation}

This implies that each sampling step can be independently optimized given $\partial \mathbf{x}_0$, allowing us to sample a subset of sampling steps for training. 
For each optimization step, we sample $K$ equally spaced sampling steps $T_{train} := \{t_s, t_s + \lfloor \frac{T}{K} \rfloor, ..., t_s + K \lfloor \frac{T}{K} \rfloor\}$, where $K$ is the number of training steps, and $t_s$ is a random start timestep that ensures $T_{train} \subseteq {1, ..., T}$. Only the sampling steps in $T_{train}$ are trained, which reduces the computation and memory overhead and leads to faster convergence. We provide the pseudocode of DRTune in Algorithm~\ref{alg:DRTune} and highlight the differences between our approach and relevant algorithms.

\begin{algorithm}
\caption{DRTune}\label{alg:DRTune}
\begin{algorithmic}[1]
\Statex \textbf{Input:} pre-trained diffusion model weights $\theta$, reward $r$, number of training timesteps $K$, range of early stop timestep $m$. $sg$ stands for the stop gradient operation.
\While{not converged}
\State 
$
t_{train} = \left\{ 
\begin{array}{ll}
  \{1, ..., \mathrm{K}\}   &  \mathbf{if}\quad \text{DRaFT-$K$} \\
  \{i\}_{i \ge randint(1, T)}   &  \mathbf{if}\quad \text{AlignProp} \\
\end{array}
\right.
$
\If{\textcolor{blue}{DRTune}}
    \State \textcolor{green}{\texttt{\# Equally spaced timesteps.}}
    \State \textcolor{blue}{$s = randint(1, T - K \lfloor \frac{T}{K}\rfloor)$}
    \State \textcolor{blue}{$t_{train} = \{s + i \lfloor \frac{T}{K}\rfloor \mid i = 0, 1, ..., K-1\}$}
\EndIf

\If{\textcolor{orange}{ReFL} or \textcolor{blue}{DRTune}}
    \State $t_{min} = randint(1,m)$
\Else
    \State $t_{min} = 0$
\EndIf

\State $\mathbf{x}_T \sim \mathcal{N}(0, \mathbf{I})$
\For {$t = T, ..., 1$}

\If{\textcolor{blue}{DRTune}}
\State $\hat{\epsilon} = \epsilon_\theta(\textcolor{blue}{sg(\mathbf{x}_t)}, t)$
\Else
\State $\hat{\epsilon} = \epsilon_\theta(\mathbf{x}_t, t)$
\EndIf

\If{$t \notin t_{train}$}
\State $\hat{\epsilon} = sg(\hat{\epsilon})$
\EndIf

\If{$t == t_{min} $}
\State $\mathbf{x}_0 \approx intermediate\_prediction (\mathbf{x}_t, \hat{\epsilon})$ 
\State break
\EndIf

\State $\mathbf{x}_{t-1} = a_t \mathbf{x}_{t} + b_t \hat{\epsilon} + c_t \epsilon$
\EndFor
\State $\mathbf{g} = \nabla_\theta r(\mathbf{x}_0, c)$
\State $\theta \leftarrow \theta - \eta \mathbf{g}$
\EndWhile

\end{algorithmic}
\end{algorithm}

\begin{figure*}
  \centering
  \includegraphics[width=1.0\linewidth]{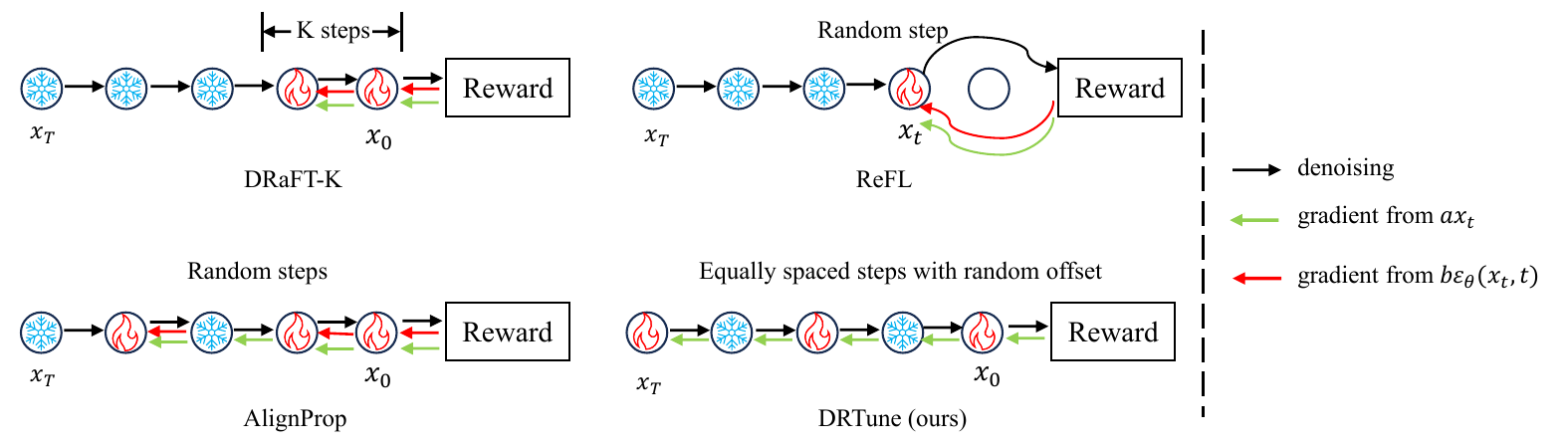}
  \caption{Comparison between reward training algorithms.}
  \label{fig:algorithm_comparison}
\end{figure*}

\noindent\textbf{Comparison with other algorithms.}
The comparison between algorithms is shown in Fig.~\ref{fig:algorithm_comparison}.
The key difference between DRTune and other algorithms lies in the stop gradient operation. We will demonstrate that other algorithms can also benefit from this simple technique in the experiment section. While all the aforementioned algorithms train a subset of steps, these steps are sampled differently. 
In ReFL~\cite{xu2023imagereward}, the step is randomly sampled near the output. The denoising loop terminates at this step, and then uses the intermediate result at that step for supervision. DRaFT-$K$~\cite{clark2023directly} only trains the last $K$ steps near the output image, without involving earlier sampling steps. AlignProp~\cite{prabhudesai2023aligning} trains a random subset of sampling steps, with a higher probability for steps near the output image and a lower probability for steps near the input noise. DRTune also trains a subset of $K$ steps, but the selected steps are equally spaced. We find that this design results in faster convergence than randomly sampling $K$ steps for training. We also adopt the early stop design of ReFL, which accelerates sampling, and leads to faster convergence.

\section{Experiments}
We present a comparison between DRTune and other reward-training methods in Sec.\ref{sec:baseline}. Following that, we analyze the design choices of DRTune through ablation experiments in Sec.\ref{sec:ablation}. Finally, we demonstrate the effectiveness of DRTune in tuning the larger model of SDXL 1.0 in Sec.\ref{sec:sdxl}.

\subsection{Comparison with baseline methods}
\label{sec:baseline}

We experiment with 7 different reward functions in our study: Aesthetic Score~\cite{ava, aesthetic_classifier}, CLIPScore~\cite{clip, openclip}, PickScore~\cite{kirstain2023pick}, HPS v2.1~\cite{wu2023humanv2}, symmetry, compressibility, and objectness.

\noindent\textbf{Aesthetic Score} is evaluated using an aesthetic score predictor~\cite{aesthetic_classifier}, which is a single-modal reward model that takes an image as input and predicts a score from 1 to 10, evaluating its aesthetic quality. This predictor is trained on a scoring dataset~\cite{ava} consisting of real images.

\noindent\textbf{CLIPScore~\cite{clip, openclip}} is defined as the cosine similarity between the text embedding of the input prompt and the image embedding of the output image. 
We use the ViT-H-14 variant of the CLIP model for evaluation.

\noindent\textbf{PickScore~\cite{kirstain2023pick}, HPS v2.1~\cite{wu2023humanv2} } are models trained to capture human preference for images based on input prompts. Both models are fine-tuned on top of the CLIP model~\cite{clip} but with different data.

\noindent\textbf{Symmetry} is a low-level reward that is defined in the pixel space. The reward function for symmetry is given by:
\begin{equation}
r_{symmetry}(I) = \frac{||I - \text{flip}(I)||_1}{\text{std}(I)},
\end{equation}
where $\text{flip}()$ applies a mirror flip to the input image, and $\text{std}()$ computes the standard deviation of pixel values in the image, which helps prevent the image from degenerating into a solid color image.
To maintain the controllability of text prompts, we also incorporate the CLIP Score~\cite{clip, openclip} as an additional regularization term.

\noindent\textbf{Compressibility}. Inspired by ~\cite{black2023training, clark2023directly}, we use the reconstruction error of images before and after JPEG compression to measure the compressibility of an image, where the error is defined as:
\begin{equation}
  e_{compress}(I) = ||I - d(c(I))||^2_2,
\end{equation}
where $d$ and $c$ are implemented as differentiable operators.

\noindent\textbf{Objectness}. Inspired by ~\cite{prabhudesai2023aligning}, we minimize the objectness score of a type of object in an image to achieve object removal.
We use OWL-ViT as the object detector, and try to remove ``books'' in images generated by prompts in the form of ``[concept name], and books.''.
In practice, we minimize the maximum objectness of class ``book'' in all the box predictions.

\noindent\textbf{Experiment setup.}
To ensure a fair comparison, we adopt a unified hyperparameter setting for all baselines. However, since each method has different GPU consumption, comparing them based on the same training steps or reward query count would be unfair for faster methods. Instead, we evaluate the methods using the same computation budget. Specifically, we train all methods for 2 hours on 4 A100 GPUs. The final reward is computed on the 400 prompts from the HPS v2 benchmark~\cite{wu2023humanv2}.
Since this work focuses on optimizing a given reward, common evaluation metrics for image creation, like FiD~\cite{fid} and IS~\cite{inceptionscore} are not reported.

\noindent\textbf{Implementation details.}
\label{sec:implementation}
We use Stable Diffusion 1.5~\cite{stable_diffusion} as the base model for our experiments. We employ a DDPM~\cite{ho2020denoising} sampler to perform 50 steps of sampling, with a classifier-free guidance scale of 7.5. The output resolution of the generated images is set to $512\times512$, which is then down-sampled to $224\times224$ for reward evaluation.
During training, we sample prompts from the training prompts provided by HPS v2~\cite{wu2023humanv2}. The models are trained with a batch size of 32 and a constant learning rate of $2\times10^{-5}$. We apply gradient clipping of 0.1 to ensure stable training. The AdamW optimizer with default hyperparameters ($\beta_1 = 0.9, \beta_2 = 0.999$)~\cite{adam, adamw} is used for parameter optimization.
To save memory during training, we train LoRA~\cite{hu2021lora} weights of rank 128 instead of the entire network. Additionally, we use half-precision (fp16) for frozen parameters and single precision for trainable parameters. Gradient checkpointing is also applied to further reduce memory usage.
It is worth noting that AlignProp~\cite{prabhudesai2023aligning} requires significantly more memory for optimization. Therefore, in order to make a fair comparison with other methods, we perform gradient accumulation of 2 steps in AlignProp.
The Stable Diffusion 1.5 model consists of three block modules: the VAE, the U-Net, and the text encoder. In our training process, we only train the LoRA parameters in the U-Net module.

\begin{figure*}
  \centering
    \includegraphics[width=1.0\linewidth]{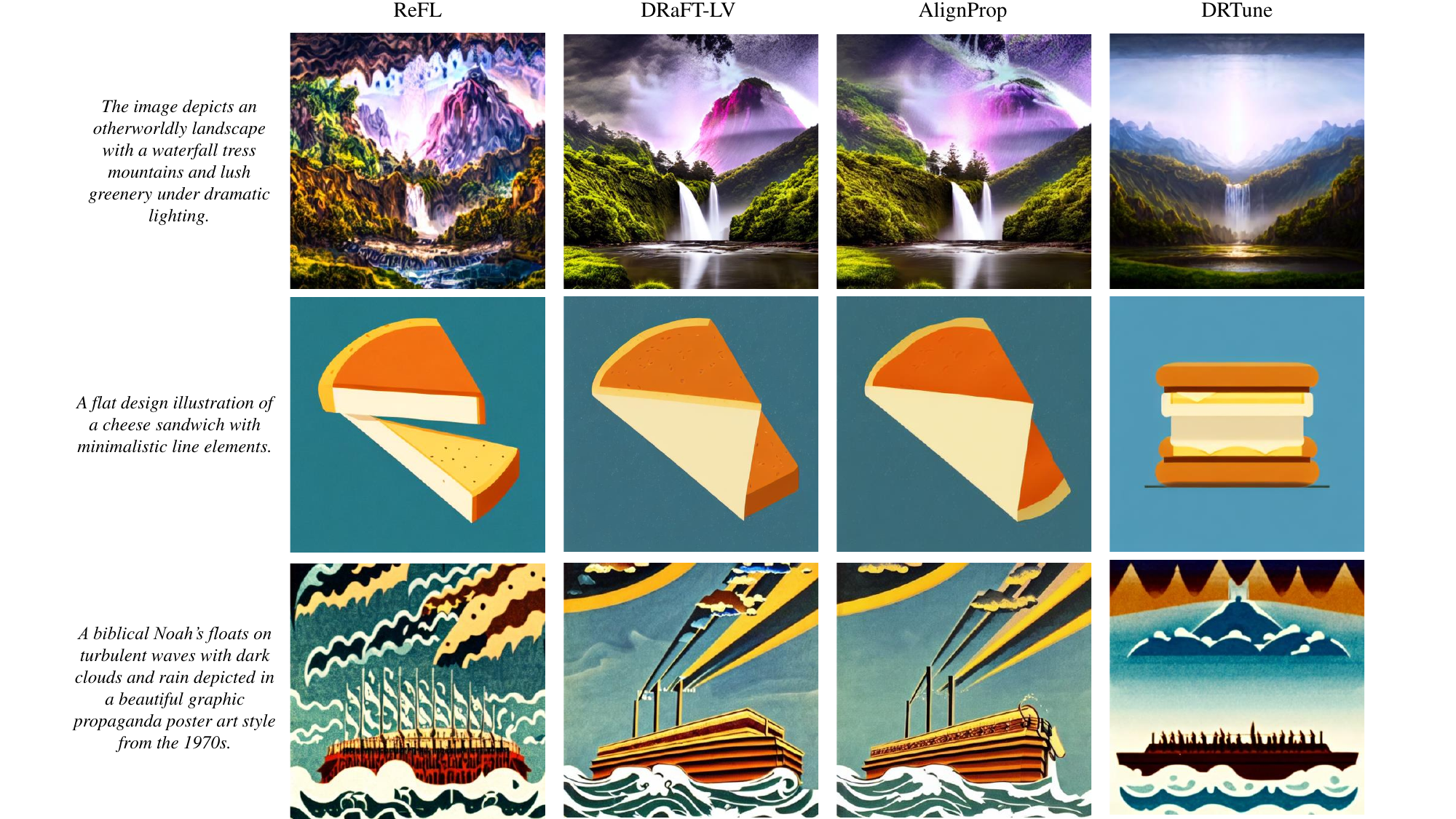}

  \caption{Comparison between images generated with DRTune or other baseline methods and {\bf symmetry} supervision.}
  \label{fig:compare_symmetry}
\end{figure*}

\begin{figure*}
  \centering
    \includegraphics[width=1.0\linewidth]{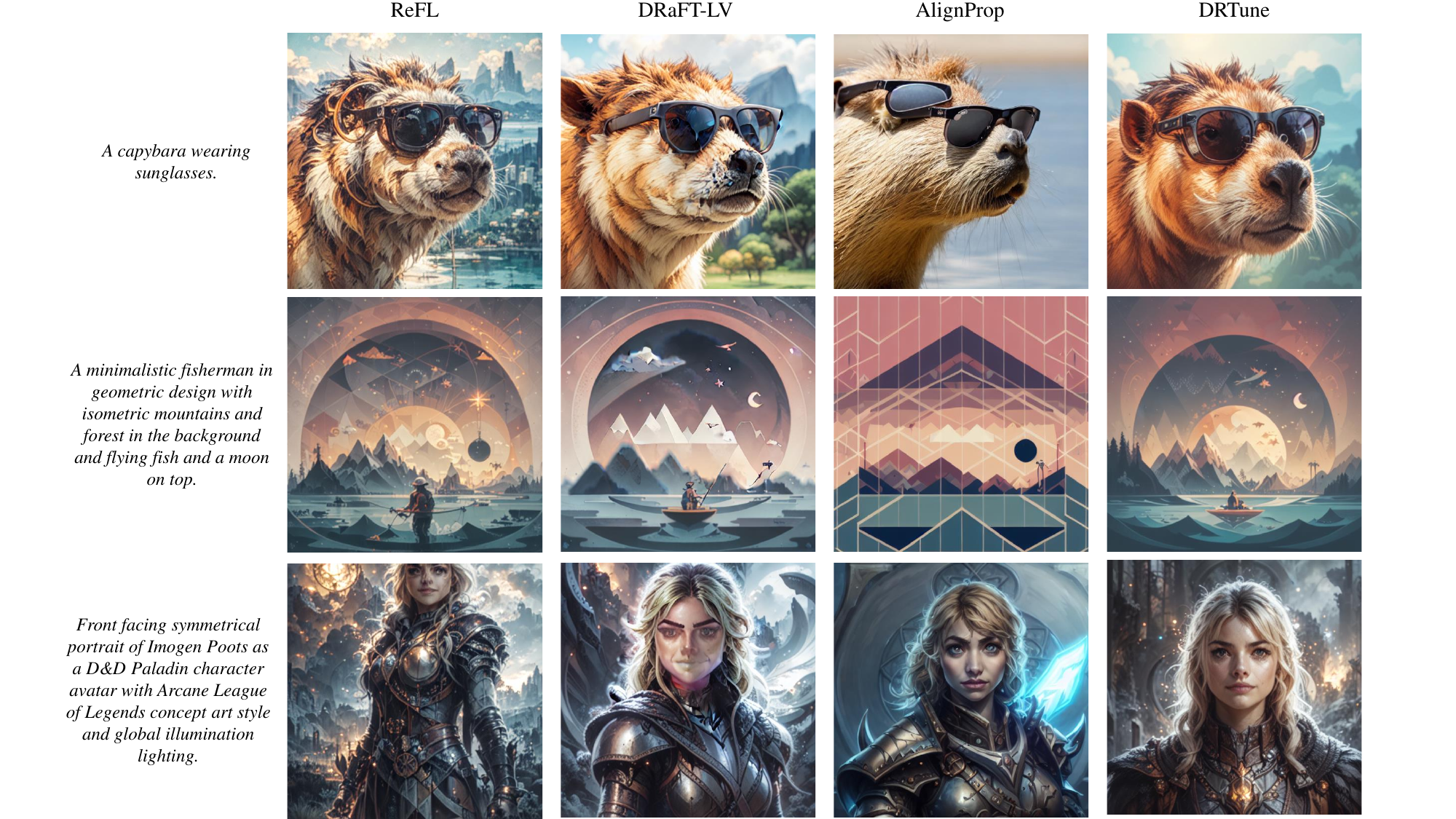}

  \caption{Comparison between images generated with Stable Diffusion tuned by DRTune or other baseline methods. Supervised by PickScore.}
  \label{fig:compare_pickscore}
\end{figure*}

\noindent\textbf{Comparison between reward-training methods.}
In Tab.~\ref{tab:main_result}, we compare DRTune with variants of each baseline. ReFL-10 and ReFL-20~\cite{xu2023imagereward} are variants of ReFL that stops at different range of steps. 
ReFL runs faster than other baselines due to its early stop design.
DRaFT-LV trains only the last step, but increases its efficiency by sampling the last step twice with different noise.
DRaFT-10 and AlignProp train more denoising steps in each parameter update, but they are considerably slower compared to the previous three baselines, hindering their convergence.
In contrast, DRTune samples all denoising steps without early termination. It trains 5 denoising steps and achieves the highest efficiency among the baselines. An important feature of DRTune is its ability to successfully optimize the symmetry loss, which sets it apart from the other algorithms. This success can be attributed to DRTune's efficient supervision of early denoising steps, which is better related to the global layout of an image.
In Fig.~\ref{fig:compare_symmetry} and Fig.~\ref{fig:compare_pickscore}, we show qualitative comparison between methods on the reward of PickScore and symmetry. Our method successfully learns the symmetry property, while others fail. And with supervision on the early denoising steps, models tuned with DRTune often shows better global layout and more natural appearance than other methods.

\begin{table}
  \centering
  \resizebox{1.0\textwidth}{!}{
  \begin{tabular}{lcccccccc}
    \toprule
    Method & HPS v2.1$\uparrow$ & PickScore$\uparrow$ & Aesthetic Score$\uparrow$ & Symmetry$\downarrow$ & CLIPScore $\uparrow$  &  Compression error $\downarrow$ & Objectness $\downarrow$ \\
    \midrule
    ReFL-10~\cite{xu2023imagereward} & 39.18 & 0.2488 & 9.585 & 0.7512 & 0.3771 & 0.0057 & 0.0029\\
    ReFL-20~\cite{xu2023imagereward} & 39.43 & 0.2491 & 9.911 & 0.7328 & 0.3844 & 0.0043 & 0.0021\\ 
    DRaFT-LV~\cite{clark2023directly} & 39.02 & 0.2477 & 9.615 & 0.7648 & 0.3613 & 0.0088 & 0.1024\\
    DRaFT-10~\cite{clark2023directly} & 37.62 & 0.2456 & 8.317 & 0.7250  & 0.3707 & 0.0073 & 0.0050\\
    AlignProp~\cite{prabhudesai2023aligning} & 34.04 & 0.2297 & 6.627 & 0.7840 & 0.3580 & 0.0097 & 0.9266 \\
    \midrule
    DRTune (outs) & \textbf{40.63} & \textbf{0.2492} & \textbf{10.360} & \textbf{0.0551} & \textbf{0.3856} & \textbf{0.0038} & \textbf{0.0001}\\
    \bottomrule
  \end{tabular}
  }
  \caption{Comparison with baseline methods. The diffusion model is trained to maximize or minimize the given targets, indicated by top / down arrows. We report averaged targets computed on prompts from the HPS v2~\cite{wu2023humanv2} benchmark except for the objectness. Prompts for objectness are in the form of ``[concept name], and books'', following ~\cite{prabhudesai2023aligning}.}
  \label{tab:main_result}
\end{table}

\begin{figure}
  \centering
  \begin{subfigure}{0.7\linewidth}
    \includegraphics[width=1.0\linewidth]{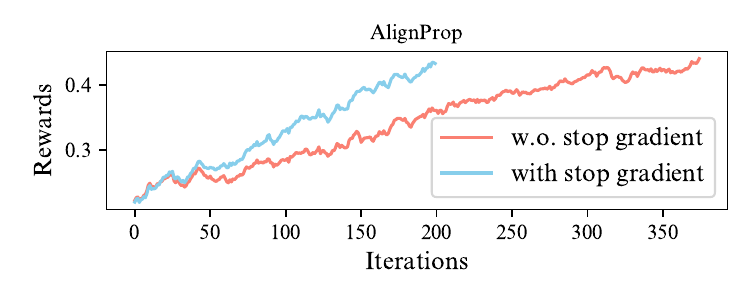}
    \label{fig:short-a}
  \end{subfigure}
  \hfill
  \begin{subfigure}{0.7\linewidth}
    \includegraphics[width=1.0\linewidth]{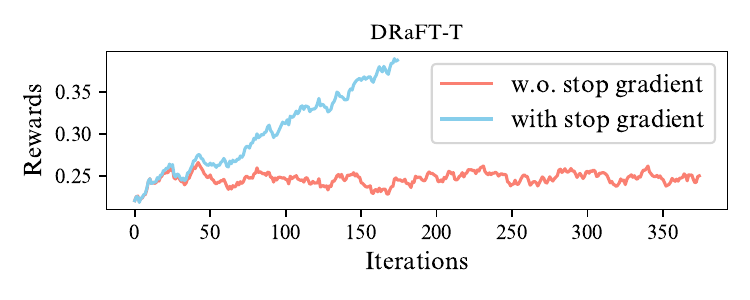}
    \label{fig:short-b}
  \end{subfigure}
  \caption{AlignProp~\cite{prabhudesai2023aligning} and DRaFT-$T$~\cite{clark2023directly} can also benefit from early stop.}
  \label{fig:stop_gradient}
\end{figure}

\begin{figure}
  \centering
  \includegraphics[width=0.7\linewidth]{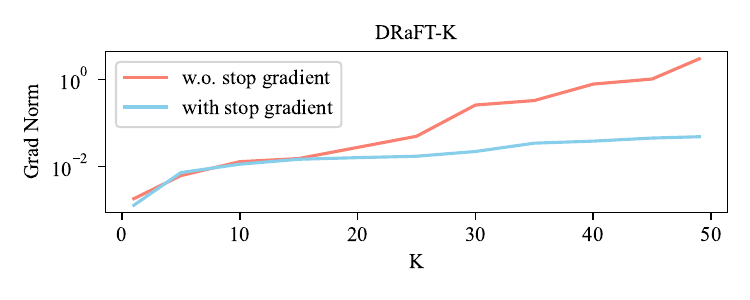}
  \caption{The gradient norm of network parameters explodes as $K$ increases, which can be mitigated by the stop gradient technique. }
  \label{fig:grad_norm}
\end{figure}

\begin{figure}
  \centering
  \includegraphics[width=0.7\linewidth]{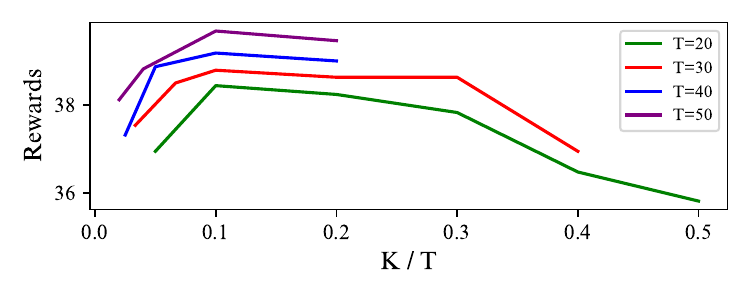}
  \caption{The impact of the number of training steps $K$ vs. various total sampling steps $T$. Some data points are missing due to GPU memory constraint.}
  \label{fig:K_ablation}
\end{figure}

\subsection{Ablation studies}
\label{sec:ablation}
\noindent\textbf{Stop gradient for deep supervision.}
Stop gradient is a general reward-training technique that can greatly help the convergence when training diffusion models with a reward. In Fig.~\ref{fig:stop_gradient}, we present the convergence curves of DRaFT-$T$\cite{clark2023directly} and AlignProp~\cite{prabhudesai2023aligning} trained with and without the proposed stop gradient technique.
Both algorithms involve supervision on early denoising steps.
DRaFT-$T$ is a variant of DRaFT-$K$ that trains all the $T$ denoising steps, while AlignProp trains $\frac{T}{2}$ steps on average. DRaFT-$T$ fails to optimize without the stop gradient technique.
In the case of AlignProp, the stop gradient technique significantly improves the training efficiency, as it only requires roughly 50\% of iterations to achieve a similar reward. The effectiveness of the stop gradient technique can be explained by the gradient norm. In Fig.~\ref{fig:grad_norm}, we plot the gradient norm of LoRA parameters of the U-Net module. Without the stop gradient technique, the gradient norm rapidly increases after $K\ge15$. However, this issue is effectively resolved by adopting the proposed stop gradient technique.

\noindent\textbf{Choice of K.}
$K$ is the number of denoising steps to be trained in each iteration. While a larger $K$ value requires more computational resources for back-propagation, it may also help to acquire more accurate gradient estimation. In Fig.~\ref{fig:K_ablation}, we conduct an ablation study on the selection of $K$ in DRTune, considering different total sampling steps $T$ ranging from 20 to 50.
To ensure fair comparison, experiments with the same $T$ were conducted under identical computational budget and hyperparameters. 
The final reward is plotted against the ratio of $\frac{K}{T}$.
The results demonstrate that the optimal value of $K$ exhibits a linear relationship with $T$, and a ratio of $0.1T$ turns out to be a favorable choice.

\begin{figure}
  \centering
  \includegraphics[width=0.6\linewidth]{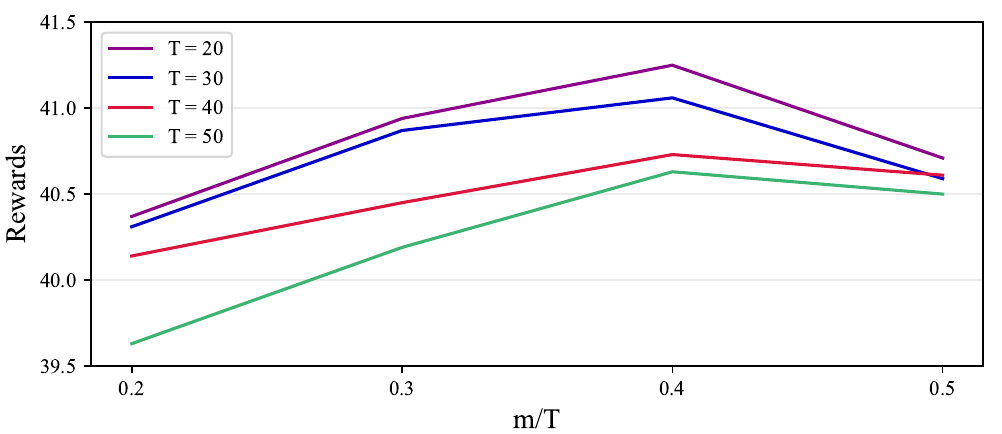}
  \caption{The impact of maximal early stop step $m$ vs. various total sampling steps $T$.}
  \label{fig:ablate_m}
\end{figure}

\noindent\textbf{Choice of m.}
$m$ controls early stop, which is the maximal number of steps to skip. In Fig.~\ref{fig:ablate_m}, we ablate the choice of $m$. The result shows that the best choice of $m$ falls on the ratio of $0.4T$.



\begin{figure*}
  \centering
  \includegraphics[width=0.9\linewidth]{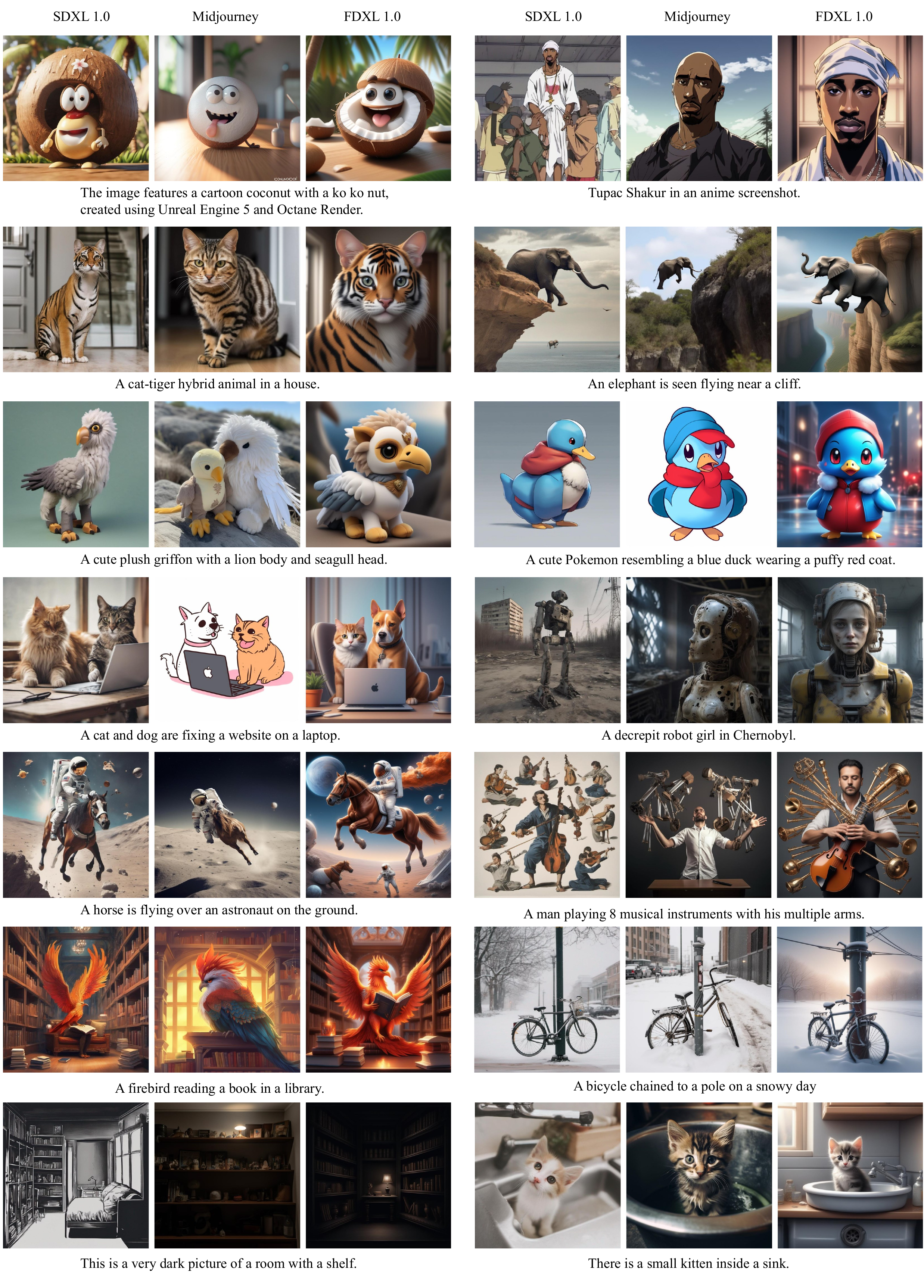}
  \caption{Qualitative comparison between SDXL, Midjourney and FDXL.}
  \label{fig:fdxl_comparison}
\end{figure*}

\subsection{Training SDXL 1.0}
\label{sec:sdxl}
\noindent\textbf{Training details.} 
We fine-tune the SDXL 1.0 model to optimize HPS v2.1 using DRTune, resulting in Favorable Diffusion XL 1.0 (FDXL 1.0). Given SDXL's larger size compared to Stable Diffusion 1.5, we utilize the DPM++\cite{dpm++} sampler for improved sampling efficiency. We adjust the sampling steps $T$ to 25 and the training step $K$ to 3, maintaining a classifier-free guidance scale of 5.0 and a resolution of 1024 to align with SDXL 1.0's default setting. We train the model using a batch size of 2 and a gradient accumulation step of 4 on 8 A100 GPUs with 80G memory for 1,900 iterations. We lower the learning rate to $5\times10^{-6}$ for training stability. To address the discrepancy between the input resolution of the reward model and the output image resolution of SDXL, we apply random shifting within the range of $\{0, ..., \lceil \frac{r_{\text{output}}}{r_{\text{input}}}\rceil \}$ pixels to the output image, randomizing the down-sampling pixel locations. We set the LoRA weight to 0.7 during inference. All other settings remain consistent with those outlined in Sec.~\ref{sec:implementation}.

\begin{figure}[tb]
  \centering
  \begin{subfigure}{0.49\linewidth}
  \includegraphics[width=1.0\linewidth]{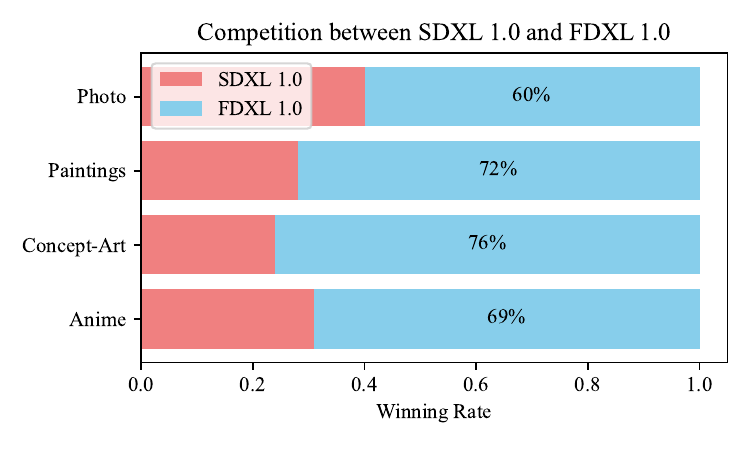}
  \caption{User study comparing images generated by SDXL 1.0 and FDXL 1.0.}
  \label{fig:vs_sdxl}
  \end{subfigure}
  \hfill
  \begin{subfigure}{0.49\linewidth}
  \includegraphics[width=1.0\linewidth]{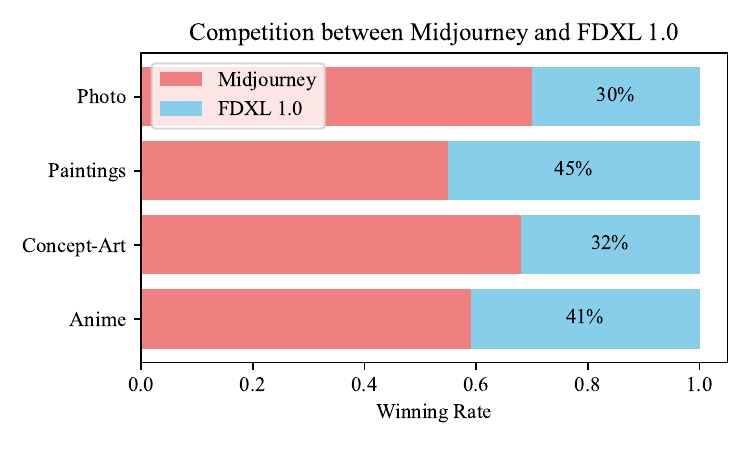}
  \caption{User study comparing images generated by Midjourney and FDXL 1.0.}
  \label{fig:vs_mj}
  \end{subfigure}
\end{figure}



\subsection{Results}
We compare images generated by FDXL 1.0 with SDXL 1.0 and the commercial generative model Midjourney. In Fig.\ref{fig:fdxl_comparison}, we present the visual comparison between images generated by SDXL, Midjourney, and FDXL. FDXL produces higher-quality and more contextually consistent images than SDXL. For quantitative comparison, we conduct user studies on 400 randomly sampled prompts from the HPS v2\cite{wu2023humanv2} benchmark, with 100 prompts from each category of ``Photo'', ``Paintings'', ``Concept-Art'', and ``Anime''. Each prompt is evaluated by 3 participants, and the results are averaged for visualization, as depicted in Fig.\ref{fig:vs_sdxl}, FDXL significantly outperforms SDXL, with an average winning rate of 69\%. In Fig.\ref{fig:vs_mj}, we illustrate the comparison between FDXL and Midjourney, showing an average winning rate of 37\% compared to Midjourney, with the ``paintings'' category approaching a draw. However, we acknowledge that the realism of images generated by FDXL still lags behind Midjourney, and the winning rate over SDXL on prompts of ``photo'' category is also the lowest among others.
AS this is potentially due to bias in the reward model, the challenge of leveraging a preference model while mitigating its bias still remains an open question.

\section{Limitations}
In this work, we primarily focus on optimizing a given reward, but we find in our experiments that rewards that are computed from a neural network are very likely to suffer from the reward-hacking problem. 
For example, an image generated by an over-optimized model can have a very high aesthetic score, but the visual quality may degenerate. 
When training FDXL, we avoid the issue by early stop. It is also meaningful to explore other strategies like regularization.

Generative models potentially have negative social impacts by their nature. The advancement in generative models often means more plausible generated content, which may be maliciously utilized for the spread of more convincing misinformation and fake content. Additionally, there is a risk of amplifying existing biases and stereotypes present in the training data.

\section{Conclusion}
In this study, we address the challenge of training text-to-image diffusion models using feedback from a reward model. We emphasize the importance of deep supervision for optimizing global rewards and resolve convergence issues using a stop gradient technique. Additionally, we demonstrate the potential of reward training by fine-tuning the FDXL 1.0 model to achieve image quality comparable to Midjourney.

\clearpage

%
%
\bibliographystyle{splncs04}
\bibliography{main}

\begin{thebibliography}{10}
\providecommand{\url}[1]{\texttt{#1}}
\providecommand{\urlprefix}{URL }
\providecommand{\doi}[1]{https://doi.org/#1}

\bibitem{bai2022training}
Bai, Y., Jones, A., Ndousse, K., Askell, A., Chen, A., DasSarma, N., Drain, D., Fort, S., Ganguli, D., Henighan, T., et~al.: Training a helpful and harmless assistant with reinforcement learning from human feedback. arXiv preprint arXiv:2204.05862  (2022)

\bibitem{black2023training}
Black, K., Janner, M., Du, Y., Kostrikov, I., Levine, S.: {Training diffusion models with reinforcement learning}. arXiv preprint arXiv:2305.13301  (2023)

\bibitem{clark2023directly}
Clark, K., Vicol, P., Swersky, K., Fleet, D.J.: {Directly Fine-Tuning Diffusion Models on Differentiable Rewards}. arXiv preprint arXiv:2309.17400  (2023)

\bibitem{fan2023dpok}
Fan, Y., Watkins, O., Du, Y., Liu, H., Ryu, M., Boutilier, C., Abbeel, P., Ghavamzadeh, M., Lee, K., Lee, K.: {DPOK: Reinforcement Learning for Fine-tuning Text-to-Image Diffusion Models}. arXiv preprint arXiv:2305.16381  (2023)

\bibitem{goodfellow2020generative}
Goodfellow, I., Pouget-Abadie, J., Mirza, M., Xu, B., Warde-Farley, D., Ozair, S., Courville, A., Bengio, Y.: Generative adversarial networks. Communications of the ACM  \textbf{63}(11),  139--144 (2020)

\bibitem{fid}
Heusel, M., Ramsauer, H., Unterthiner, T., Nessler, B., Hochreiter, S.: {GANs Trained by a Two Time-Scale Update Rule Converge to a Local Nash Equilibrium}. In: NeurIPS (2017)

\bibitem{ho2020denoising}
Ho, J., Jain, A., Abbeel, P.: {Denoising diffusion probabilistic models}. NeurIPS  \textbf{33},  6840--6851 (2020)

\bibitem{hu2021lora}
Hu, E.J., Shen, Y., Wallis, P., Allen-Zhu, Z., Li, Y., Wang, S., Wang, L., Chen, W.: {Lora: Low-rank adaptation of large language models}. arXiv preprint arXiv:2106.09685  (2021)

\bibitem{openclip}
Ilharco, G., Wortsman, M., Wightman, R., Gordon, C., Carlini, N., Taori, R., Dave, A., Shankar, V., Namkoong, H., Miller, J., Hajishirzi, H., Farhadi, A., Schmidt, L.: Openclip (Jul 2021). \doi{10.5281/zenodo.5143773}, \url{https://doi.org/10.5281/zenodo.5143773}, if you use this software, please cite it as below.

\bibitem{kim2022diffusionclip}
Kim, G., Kwon, T., Ye, J.C.: {Diffusionclip: Text-guided diffusion models for robust image manipulation}. In: Proceedings of the IEEE/CVF Conference on Computer Vision and Pattern Recognition. pp. 2426--2435 (2022)

\bibitem{adam}
Kingma, D.P., Ba, J.: {Adam: A method for stochastic optimization}. arXiv preprint arXiv:1412.6980  (2014)

\bibitem{kirstain2023pick}
Kirstain, Y., Polyak, A., Singer, U., Matiana, S., Penna, J., Levy, O.: {Pick-a-Pic: An Open Dataset of User Preferences for Text-to-Image Generation}. arXiv preprint arXiv:2305.01569  (2023)

\bibitem{lee2023aligning}
Lee, K., Liu, H., Ryu, M., Watkins, O., Du, Y., Boutilier, C., Abbeel, P., Ghavamzadeh, M., Gu, S.S.: {Aligning text-to-image models using human feedback}. arXiv preprint arXiv:2302.12192  (2023)

\bibitem{adamw}
Loshchilov, I., Hutter, F.: {Decoupled weight decay regularization}. arXiv preprint arXiv:1711.05101  (2017)

\bibitem{dpm}
Lu, C., Zhou, Y., Bao, F., Chen, J., Li, C., Zhu, J.: {Dpm-solver: A fast ode solver for diffusion probabilistic model sampling in around 10 steps}. Advances in Neural Information Processing Systems  \textbf{35},  5775--5787 (2022)

\bibitem{dpm++}
Lu, C., Zhou, Y., Bao, F., Chen, J., Li, C., Zhu, J.: {Dpm-solver++: Fast solver for guided sampling of diffusion probabilistic models}. arXiv preprint arXiv:2211.01095  (2022)

\bibitem{ava}
Murray, N., Marchesotti, L., Perronnin, F.: {AVA: A large-scale database for aesthetic visual analysis}. CVPR pp. 2408--2415 (2012)

\bibitem{ouyang2022training}
Ouyang, L., Wu, J., Jiang, X., Almeida, D., Wainwright, C., Mishkin, P., Zhang, C., Agarwal, S., Slama, K., Ray, A., et~al.: {Training language models to follow instructions with human feedback}. NeurIPS  \textbf{35},  27730--27744 (2022)

\bibitem{piao2021inverting}
Piao, J., Sun, K., Wang, Q., Lin, K.Y., Li, H.: Inverting generative adversarial renderer for face reconstruction. In: Proceedings of the IEEE/CVF Conference on Computer Vision and Pattern Recognition. pp. 15619--15628 (2021)

\bibitem{podell2023sdxl}
Podell, D., English, Z., Lacey, K., Blattmann, A., Dockhorn, T., M{\"u}ller, J., Penna, J., Rombach, R.: Sdxl: Improving latent diffusion models for high-resolution image synthesis. arXiv preprint arXiv:2307.01952  (2023)

\bibitem{prabhudesai2023aligning}
Prabhudesai, M., Goyal, A., Pathak, D., Fragkiadaki, K.: {Aligning Text-to-Image Diffusion Models with Reward Backpropagation}. arXiv preprint arXiv:2310.03739  (2023)

\bibitem{clip}
Radford, A., Kim, J.W., Hallacy, C., Ramesh, A., Goh, G., Agarwal, S., Sastry, G., Askell, A., Mishkin, P., Clark, J., Krueger, G., Sutskever, I.: {Learning Transferable Visual Models From Natural Language Supervision}. In: ICML (2021)

\bibitem{dalle2}
Ramesh, A., Dhariwal, P., Nichol, A., Chu, C., Chen, M.: {Hierarchical Text-Conditional Image Generation with CLIP Latents}. ArXiv  \textbf{abs/2204.06125} (2022)

\bibitem{ren2015faster}
Ren, S., He, K., Girshick, R., Sun, J.: Faster r-cnn: Towards real-time object detection with region proposal networks. Advances in neural information processing systems  \textbf{28} (2015)

\bibitem{stable_diffusion}
Rombach, R., Blattmann, A., Lorenz, D., Esser, P., Ommer, B.: {High-Resolution Image Synthesis with Latent Diffusion Models}. CVPR pp. 10674--10685 (2022)

\bibitem{ruiz2023dreambooth}
Ruiz, N., Li, Y., Jampani, V., Pritch, Y., Rubinstein, M., Aberman, K.: {Dreambooth: Fine tuning text-to-image diffusion models for subject-driven generation}. In: Proceedings of the IEEE/CVF Conference on Computer Vision and Pattern Recognition. pp. 22500--22510 (2023)

\bibitem{imagen}
Saharia, C., Chan, W., Saxena, S., Li, L., Whang, J., Denton, E.L., Ghasemipour, K., Gontijo~Lopes, R., Karagol~Ayan, B., Salimans, T., et~al.: {Photorealistic text-to-image diffusion models with deep language understanding}. NeurIPS  \textbf{35},  36479--36494 (2022)

\bibitem{inceptionscore}
Salimans, T., Goodfellow, I., Zaremba, W., Cheung, V., Radford, A., Chen, X.: {Improved techniques for training gans}. Advances in neural information processing systems  \textbf{29} (2016)

\bibitem{aesthetic_classifier}
Schuhmann, C.: {CLIP+MLP Aesthetic Score Predictor}. \url{https://github.com/christophschuhmann/improved-aesthetic-predictor} (2022)

\bibitem{sohl2015deep}
Sohl-Dickstein, J., Weiss, E., Maheswaranathan, N., Ganguli, S.: {Deep unsupervised learning using nonequilibrium thermodynamics}. In: International conference on machine learning. pp. 2256--2265. PMLR (2015)

\bibitem{ddim}
Song, J., Meng, C., Ermon, S.: {Denoising diffusion implicit models}. arXiv preprint arXiv:2010.02502  (2020)

\bibitem{stiennon2020learning}
Stiennon, N., Ouyang, L., Wu, J., Ziegler, D., Lowe, R., Voss, C., Radford, A., Amodei, D., Christiano, P.F.: Learning to summarize with human feedback. Advances in Neural Information Processing Systems  \textbf{33},  3008--3021 (2020)

\bibitem{sun2022controllable}
Sun, K., Wu, S., Huang, Z., Zhang, N., Wang, Q., Li, H.: Controllable 3d face synthesis with conditional generative occupancy fields. Advances in Neural Information Processing Systems  \textbf{35},  16331--16343 (2022)

\bibitem{sun2022cgof++}
Sun, K., Wu, S., Zhang, N., Huang, Z., Wang, Q., Li, H.: Cgof++: Controllable 3d face synthesis with conditional generative occupancy fields. IEEE transactions on pattern analysis and machine intelligence  (2023)

\bibitem{teed2020raft}
Teed, Z., Deng, J.: {Raft: Recurrent all-pairs field transforms for optical flow}. In: Computer Vision--ECCV 2020: 16th European Conference, Glasgow, UK, August 23--28, 2020, Proceedings, Part II 16. pp. 402--419. Springer (2020)

\bibitem{wallace2023end}
Wallace, B., Gokul, A., Ermon, S., Naik, N.: End-to-end diffusion latent optimization improves classifier guidance. arXiv preprint arXiv:2303.13703  (2023)

\bibitem{watson2021learning}
Watson, D., Chan, W., Ho, J., Norouzi, M.: Learning fast samplers for diffusion models by differentiating through sample quality. In: International Conference on Learning Representations (2021)

\bibitem{wu2023humanv2}
Wu, X., Hao, Y., Sun, K., Chen, Y., Zhu, F., Zhao, R., Li, H.: Human preference score v2: A solid benchmark for evaluating human preferences of text-to-image synthesis. arXiv preprint arXiv:2306.09341  (2023)

\bibitem{wu2023better}
Wu, X., Sun, K., Zhu, F., Zhao, R., Li, H.: Better aligning text-to-image models with human preference. arXiv preprint arXiv:2303.14420  (2023)

\bibitem{wu2023human}
Wu, X., Sun, K., Zhu, F., Zhao, R., Li, H.: {Human Preference Score: Better Aligning Text-to-Image Models with Human Preference}. In: Proceedings of the IEEE/CVF International Conference on Computer Vision. pp. 2096--2105 (2023)

\bibitem{xu2023imagereward}
Xu, J., Liu, X., Wu, Y., Tong, Y., Li, Q., Ding, M., Tang, J., Dong, Y.: {ImageReward: Learning and Evaluating Human Preferences for Text-to-Image Generation} (2023)

\bibitem{zhang2022dino}
Zhang, H., Li, F., Liu, S., Zhang, L., Su, H., Zhu, J., Ni, L.M., Shum, H.Y.: {Dino: Detr with improved denoising anchor boxes for end-to-end object detection}. arXiv preprint arXiv:2203.03605  (2022)

\bibitem{zhang2023adding}
Zhang, L., Rao, A., Agrawala, M.: {Adding conditional control to text-to-image diffusion models}. In: Proceedings of the IEEE/CVF International Conference on Computer Vision. pp. 3836--3847 (2023)

\bibitem{zhu2020deformable}
Zhu, X., Su, W., Lu, L., Li, B., Wang, X., Dai, J.: {Deformable detr: Deformable transformers for end-to-end object detection}. arXiv preprint arXiv:2010.04159  (2020)

\bibitem{ziegler2019fine}
Ziegler, D.M., Stiennon, N., Wu, J., Brown, T.B., Radford, A., Amodei, D., Christiano, P., Irving, G.: Fine-tuning language models from human preferences. arXiv preprint arXiv:1909.08593  (2019)

\end{thebibliography}
\end{document}


\section{Images used in user study}
Please click visualize.html in the same folder to visualize all the images in the user study. 
The first column is generated by FDXL (ours).
The second column is generated by Midjourney v5.2. 
The third column is generated by SDXL v1.0.
Prompts are listed below each row of images.

\section{Comparison with DDPO}
In Tab.~\ref{tab:ddpo}, we compare our method with DDPO~\cite{black2023training} that does not depend on gradients from the reward model.
Although DDPO can optimize non-differentiable rewards that DRTune cannot optimize, DRTune is significantly faster than DDPO.

\begin{table}
  \centering
  \resizebox{1.0\textwidth}{!}{
  \begin{tabular}{lcccccccc}
    \toprule
    Method & HPS v2.1$\uparrow$ & PickScore$\uparrow$ & Aesthetic Score$\uparrow$ & Symmetry$\downarrow$ & CLIPScore $\uparrow$  &  Compression error $\downarrow$ & Objectness $\downarrow$ \\
    \midrule
    DDPO~\cite{black2023training} & 26.48 & 0.2131 & 5.702 & 3.015 & 0.3521 & 0.0090 & 0.9544 \\
    DRTune (outs) & \textbf{40.63} & \textbf{0.2492} & \textbf{10.360} & \textbf{0.0551} & \textbf{0.3856} & \textbf{0.0038} & \textbf{0.0001}\\
    \bottomrule
  \end{tabular}
  }
  \caption{Comparison with DDPO~\cite{black2023training}. The diffusion model is trained to maximize or minimize the given targets, indicated by top / down arrows. We report averaged targets computed on prompts from the HPS v2~\cite{wu2023humanv2} benchmark except for the objectness. Prompts for objectness are in the form of ``[concept name], and books'', following ~\cite{prabhudesai2023aligning}.}
  \label{tab:ddpo}
\end{table}
\bibliographystyle{splncs04}
\bibliography{main}